\documentclass[sn-mathphys,Numbered,pdflatex]{sn-jnl}

\usepackage{graphicx}%
\usepackage{multirow}%
\usepackage{amsmath,amssymb,amsfonts}%
\usepackage{amsthm}%
\usepackage{mathrsfs}%
\usepackage[title]{appendix}%
\usepackage{xcolor}%
\usepackage{textcomp}%
\usepackage{manyfoot}%
\usepackage{booktabs}%
\usepackage{algorithm}%
\usepackage{algorithmicx}%
\usepackage{algpseudocode}%
\usepackage{listings}%

\begin{document}

\title[Article Title]{Transduce: learning transduction grammars for string transformation}

\author[1]{Francis Frydman}
\author[2]{Philippe Mangion}
\email{francis.frydman@gmail.com}

\abstract{The synthesis of string transformation programs from input-output examples utilizes various techniques, all based on an inductive bias that comprises a restricted set of basic operators to be combined. A new algorithm, Transduce, is proposed, which is founded on the construction of abstract transduction grammars and their generalization. We experimentally demonstrate that Transduce can learn positional transformations efficiently from one or two positive examples without inductive bias, achieving a success rate higher than the current state of the art.}




\maketitle
\section{Introduction}\label{sec1}

String transformation is one of the few domains where automatic program synthesis from examples (PBE) has already found commercial application, thanks to the FlashFill (Gulwani, 2011) program included in Microsoft Excel 2013 and later.

Inductive Logic Programming (ILP)  (Muggleton, 1991) can also be used to synthesize such transformation programs (Cropper, 2015).

Learning in both cases depends on an inductive bias achieved by selecting a group of basic operators that will be combined in the synthesized programs.

We propose a new algorithm, Transduce, based on the construction of abstract transduction grammars, and their generalization. 

\footnotetext[1]{No affiliation. Paris, France. ORCID:\url{https://orcid.org/0009-0001-3610-3920}}
\footnotetext[2]{No affiliation. Paris, France. Philippe Mangion provided critical proofreading and editing.}
We experimentally demonstrate that Transduce can efficiently learn positional transformations, without inductive bias, achieving a success rate significantly higher than FlashFill (88.64\% vs. 75\%), albeit requiring slightly more examples on average (Transduce: 1.41; FlashFill: 1.36).

\section{Related works }\label{sec2}

Several techniques have been proposed for generating string transformation programs. 

While some rely on the learning of regular expressions (Bartoli, 2013), others employ the learning of transduction automata (Oncina, 1993). Regardless of the approach, the limitations imposed are those of regular languages and finite stackless automata. Notably, these techniques cannot accommodate the inversion of a string, rendering it unlearnable.

The latest iteration of Inductive Logic Programming (ILP), Meta-Interpretive Learning (MIL) as executed by Metagol (Cropper, 2015), and program synthesis in dedicated languages (DSL) like the one utilized by FlashFill (Gulwani, 2011), are capable of inducing string transformations of all kinds, if the necessary operators are provided. However, an increase in the number of operators also increases the space of learnable programs, leading to longer learning time.

\section{The Transduce algorithm }\label{sec3}

\subsection{Definitions }\label{subsec1}

Let's introduce some definitions:

\textbf{Definition 1} A rational transduction is a word transformation defined by a finite-state transducer or by means of a rational relation.

 \textbf{Definition 2} A transduction grammar rule is an extension of grammar rules which, in addition to checking whether a word belongs to a rational language, as performed by a simple grammar rule, performs a given transformation on this word, producing a new word belonging to the same language.

In Transduce, words are represented by lists, and the transduction grammar rules are Prolog clauses whose leading literal is a transduction and whose body literals consist solely of the list/3 predicate defined by: list([X|Y],X,Y). which perform the transduction.

\textbf{Example 1} The transduction reverse(abc)=cba is performed by the transduction rule :

reverse([a,b,c],[c,b,a]) :- 
list([a,b,c],a,[b,c]), list([c,b,a],c,[b,a]), list([b,c],b,[c]), list([b,a],b,[a]).

 \textbf{Definition 3} An abstract transduction grammar rule is a transduction rule whose first and last arguments of all literals are variables, and whose second argument of the body literals can be a constant, a bound variable or an anonymous variable.

\subsection{How it works  }\label{subsec2}

The Transduce learning phase operates in four successive stages:
\begin{enumerate}
    \item Construction of the transduction rule for each example by successive ordered decomposition of the input and output of the examples, applying the list/3 axiom predicate.
    \item 	Abstraction of the transduction rules, by transforming each list and atomic term obtained in the previous step into a numbered variable, with the exception of constants.
    \item Encode the abstracted transduction rules as two sequences of relative integers.
    \item Compressive generalization of these two sequences by searching for number repetitions or number sequences. Two formulas are obtained, depending on the length of the example.
\end{enumerate}

The inference phase consists of two stages:
\begin{enumerate}
    \item Construction of the transduction rule corresponding to the length of the input list.
    \item Execution of the constructed rule on the input list
\end{enumerate}

\subsection{Building abstract transduction grammar rules   }\label{subsec3}

Let's take the reverse/2 predicate to be learned, with a single example:

reverse([a,b,c],[c,b,a]).\\Let the single context predicate list/3 be defined by:
 
 list([X|Y],X,Y).

\subsubsection{Construction of the basic transduction clause }\label{subsubsec2}

The list/3 predicate is iteratively applied to each argument of the predicate to be learned:

 reverse([a,b,c],[c,b,a]).

 \begin{table}[htb]
     \centering
     \begin{tabular}{|c|c|} \hline 
          [a,b,c]:& [c,b,a]:\\ \hline 
          list([a,b,c],a,[b,c])& list([c,b,a],c,[b,a]) \\ \hline 
          list([b,c],b,[c])& list([b,a],b,[a])\\ \hline
     \end{tabular}

 \end{table}

We use this decomposition to construct the C1 clause:\\
 C1:  reverse([a,b,c],[c,b,a]) :- \\list([a,b,c],a,[b,c]), list([c,b,a],c,[b,a]), list([b,c],b,[c]), list([b,a],b,[a]).\\

respecting the order of the decomposition table above (left to right and top to bottom). \\\\
The result is:\\
?- reverse([a,b,c],X).\\
X = [c,b,a].

\subsubsection{Abstracting the basic transduction clause }\label{subsubsec3}

For each list and atomic term in the C1 clause, we substitute a variable numbered in the order of appearance of the substituted term (from left to right). The result is an abstract transduction clause Cx.\\\\
Cx: reverse(X1,X2) :-\\ list(X1,X3,X4), list(X2,X5,X6), list(X4,X7,[X5]), list(X6,X7,[X3]).\\\\
with the substitutions :\\

\ [a,b,c]→X1, [c,b,a]→X2, a→X3, [b,c]→X4, c→X5, [b,a]→X6, b→X7\\\\
Finally, we eliminate the lists from the clause by introducing an intermediate variable X0 and using the standard predicate flatten/2 :\\\\
Cx': reverse(X1,X0) :- \\list(X1,X3,X4), list(X2,X5,X6), list(X4,X7,X5), list(X6,X7,X3), flatten(X2,X0).\\\\
The result is a program capable of inverting any list of exactly 3 elements:\\
?- reverse([1,2,3],X).\\
X = [3,2,1].

\subsection{Generalization of transduction clauses    }\label{subsec4}

The principle here is to :
\begin{itemize}
    \item represent the abstract clauses obtained in step 3.2.2 as sequences of numbers corresponding to the gaps between variable indices,
    \item “compress" these sequences by identifying repetitions of numbers or sequences of numbers and replacing them with a factor (number of occurrences * sequence),
    \item generalize the factors found by expressing them as a function of the length of the example.\\
\end{itemize}
If no repetition is present, there is no compression, and therefore no generalization, and Transduce asks the user for a larger example.\\
As the example reverse([a,b,c],[b,c,a]) is too short to be generalized, let's apply the algorithm from the previous section to :\\
reverse([a,b,c,d,e],[e,d,c,b,a]).\\\\
We obtain the abstract transduction clause:\\
Cx'=reverse(X1,X0) :-\\\raggedright list(X1,X3,X4), list(X2,X5,X6), list(X4,X7,X8), list(X6,X9,X10), list(X8,X11,X12), list(X10,X11,X13), list(X12,X9,X5), list(X13,X7,X3), flatten(X2,X0).\\\\
We construct the sequence of differences between the successive indices of the second arguments of the list terms:\\
X3, X5, X7, X9, X11, X11, X9, X7 → seq1=[2,2,2,2,0,-2,-2]\\\\
We compress this sequence into a formula expressed as a function of the length L of the example, which encodes sequence repetitions:\\\\
compressed seq1=(L-1)*[2],0,(L-3)*[-2]\\\\
Similarly, we construct the sequence of successive gaps between the first and third arguments of the list terms:\\
X1,X4 | X2,X6 | X4,X8 | X6,X10 | X8,X12 | X10,X13 | X12,X5 | X13,X3 \\

→ seq2= [3,4,4,4,3,-7,-10] → seq2 compressed=3,(L-2)*[4],3,-7,-10\\\\
Finally, we obtain a transduction clause with two compressed sequences (stored here in a Prolog comment):\\\\
reverse(X1,X0) :-\\\raggedright list(X1,X3,X4), list(X2,X5,X6), list(X4,X7,X8), list(X6,X9,X10), list(X8,X11,X12), list(X10,X11,X13), list(X12,X9,X5), list(X13,X7,X3), flatten(X2,X0).\\
/*suite1=(L-1)*[2],0,(L-3)*[-2] suite2=3,(L-2)*[4],3,-7,-10*/

\subsection{Inference }\label{subsec5}

When a new input of different length L is presented (execution phase), we simply calculate the sequences 1 and 2 as a function of L, which enables us to construct the corresponding transduction rule and execute it on the input.\\\\
Let's take the example above. \\\\
If we present the input :
reverse([1,2,3,4,5,6],X).\\\\
we have : L=6\\
→ seq1=5*[2],0,3*[-2] = [2,2,2,2,2,0,-2,-2,-2]\\
→ seq2=3,4*[4],3,-7,-10 = [3,4,4,4,4,3,-7,-10]\\\\
reverse(X1,X0) :-\\ \raggedright list(X1,X3,X4), list(X2,X5,X6), list(X4,X7,X8), list(X6,X9,X10), list(X8,X11,X12), list(X10,X13,X14), list(X12,X13,X15), list(X14,X11,X16), list(X15,X9,X5), list(X16,X7,X3), flatten(X2,X0).\\\\
Which results in :\\
?- reverse([a,b,c,d,e,f],X).\\
X = [f,e,d,c,b,a].

\subsection{Handling insertions and deletions  }\label{subsec6}

The algorithm described above can be used to swap elements in a list. It does not delete or add elements. To do this, we introduce the notion of transduction constants.

\subsubsection{Insertions  }\label{subsubsec4}

\textbf{Definition 4} Output transduction constant. Any element of the output argument of a transduction that is absent from the input argument.\\\\
 The abstraction step described in 3.3.2 is modified so that no variables are substituted for output constants. \\\\
 \textbf{Example 2} Consider the transduction: insert\_e([a,b,c,d],[a,b,e,c,d]). \\The element e, present in the output of the transduction and absent from its input, is an output constant of this transduction.\\\\
 The corresponding abstract transduction clause is :\\\\
insert\_e(X1,X0) :- \\
list(X1,X4,X5), list(X2,X4,X6), list(X5,X7,X8), list(X6,X7,X9), list(X8,X10,X11), list(X9,e,X8), flatten(X2,X0).\\\\
 Which yields:\\
?-  insert\_e([1,2,3,4],X).\\
X=[1,2,e,3,4].

\subsubsection{Deletions   }\label{subsubsec5}

\textbf{Definition 5} Input transduction constant. Any element of the input argument of a transduction that is absent from the output argument\\\\
The abstraction step described in 3.3.2 is modified to substitute an anonymous variable (\_) for input constants. \\\\
\textbf{Example 3} Consider the transduction: droplast([a,b,c,d],[a,b,c]). Element d, present in the input of the transduction and absent from its output, is an input constant of this transduction.\\\\
The corresponding abstract transduction clause is :\\\\
droplast(X1,X0) :- \\list(X1,X4,X5), list(X2,X4,X6), list(X5,X7,X8), list(X6,X7,X9), list(X8,X9,X10), list(X10,\_,X3), flatten(X2,X0).\\\\
This gives:\\
?- droplast([1,2,3,4],X).\\
X=[1,2,3].\\\\
\textbf{Note}: deletion is based on position alone, regardless of the value of the element being deleted.

\subsubsection{Conditional deletions    }\label{subsubsec6}

In cases where we want to condition the deletion of an element on its value, we give two example predicates in which this element has the same value.\\\\\\
 \textbf{Example 4}\\
extract([john,.,doe,@,gmail,.,com],[firstname,:,john,;,surname,:,doe]).\\
extract([mary,.,jane,@,hotmail,.,fr],[firstname,:,mary,;,surname,:,jane]).\\\\
 This results in :\\
?- extract([francis,.,frydman,@,hotmail,.,com],X)?\\
X=[firstname,:,francis,;,surname,:,frydman].\\\\
 But then :\\
?- extract([francis,dot,frydman,at,hotmail,dot,com],X)?\\
False

\subsubsection{Required number of examples }\label{subsubsec7}

 In addition to conditional deletions, some cases require two examples, for example to resolve an ambiguity. If necessary, Transduce asks for a second example of different length.\\\\
 \textbf{Example 5} \\
insert([x,y,z,t,u,v,w],[x,y,z,;,t,u,v,w]).\\\\
 Should the semicolon be inserted after the 3rd element from the beginning of the list, or before the 4th element from the end?\\\\
 The ambiguity is resolved with a second example of different length, e.g. :\\
insert([a,b,c,d,e,f,g,h],[a,b,c,;,d,e,f,g,h]).

\subsubsection{Minimum example length  }\label{subsubsec8}

When the length of the input argument of the example (or one of the examples) is insufficient to generate compressible sequences in the generalization phase (see 3.4), Transduce requests a longer example.

\section{Experiments and results  }\label{sec4}

We compare FlashFill (in Microsoft Excel) and Transduce on 44 synthetic examples (\footnotemark).
We exclude the cases involving contextual knowledge (in the form of string transformation functions, e.g. uppercase/2.) not yet processed by Transduce.

 \begin{table}[htb]
\centering

\begin{tabular}{l l l l l l l l l l}
\hline
  & Transduce &   &   &   &   &   &   & FlashFill &   \\
\hline
 \% of benchmarks solved & 88,64\% &   &  &  &  &  &   & 75,00\% &  \\
Average number of examples required & 1,41 &   &  &  &  &  &  & 1,36 &  \\
\hline

\end{tabular}

\end{table}

  Transduce's average execution time is 0.015 s per case on an Intel(R) Core(TM) i7-7700HQ 2.8 GHz, with 16 GB RAM.

\footnotetext{\url{https://github.com/ffkiller666/Transduce}}

\section{Limitations and future work  }\label{sec5}

In its current version, Transduce has limitations due to the lack of consideration of contextual knowledge, such as character transformations, which restricts the variety of cases that can be processed.

 Non-deterministic functions, which return several values, such as permutations or successive states while solving problems such as the Towers of Hanoi, are not generalized by Transduce. Generalizing such functions could lead to the synthesis of more general programs.

 Transduce only handles strings, not numbers (e.g., induction of arithmetic functions).

Additionally, the readability of induced programs is inferior to that of Metagol and FlashFill.
These issues will be addressed in future work.

 \section{Conclusion }\label{sec6}

We provided definitions for transduction grammar rules and abstract transduction grammar rules. 

Our presentation focused on a novel string transformation algorithm, Transduce, that constructs strings from input/output pairs of lists and generalizes them as two sequences of integers. Then, it generates a new transduction rule tailored to the input list's length during inference.

Finally, we have empirically shown the efficiency of Transduce by comparing it with FlashFill on a synthetic benchmark constrained to the present capacities of our algorithm. 

\section*{Declarations}
\textbf{Funding} Not Applicable.\\
\textbf{Conflicts of interest/Competing interests} The authors have no conflicts of interest to disclose.\\
\textbf{Consent for publication} Not Applicable.\\
\textbf{Consent to participate} Not Applicable.\\
\textbf{Ethics approval} Not Applicable.\\
\textbf{Code availability} The code is not publicly available.\\
\textbf{Availability of data} Data have been made available in Section 4.\\
\textbf{Authors' contributions} Author 1 wrote all sections of the paper. Author 2 provided feedback and corrections on all sections of the paper.
\section*{References}

Bartoli, A., Davanzo, G., De Lorenzo, A., Medvet, E., Sorio, E.: Automatic synthesis of regular expressions from examples. \textit{Computer}, 295–318 (2013) \url{https://doi.org/10.1109/MC.2014.344}\\\\
Cropper, A., Tamaddoni-Nezhad, A., Muggleton, S.H.: Meta-interpretive learning of data transformation programs. \textit{Inductive Logic Programming—25th International Conference, K. Inoue, H. Ohwada, A. Yamamoto (Eds.)}, 46–59 (2015) \url{https://doi.org/10.1007/978-3-319-40566-7\_4}\\\\
Gulwani, S.: Automating string processing in spreadsheets using input-output examples. \textit{POPL}, 317–330 (2011) \url{https://doi.org/10.1145/1926385.1926423}\\\\
Muggleton, S.H.: Inductive logic programming. \textit{New Generation Computing} (8), 317–330 (1991) \url{https://doi.org/10.1007/BF03037089}\\\\
Oncina, J., Garcˇıa, P., Vidal, E.: Learning subsequential transducers for pattern recognition tasks. \textit{CompIEEE Transactions on Pattern Analysis and Machine Intelligenc}e 15(5), 448–458 (1993) \url{https://doi.org/10.1109/34.211465}

\end{document}